\documentclass[runningheads]{llncs}
\usepackage{graphicx}
\usepackage{tabularx}
\usepackage{makecell} 
\usepackage{bchart}
\usepackage{pgfplots}

\usepackage{graphicx}

\usepackage{float}

\pgfplotsset{compat=1.18} 

\begin{document}
\title{Integrating HCI Datasets in Project-Based Machine Learning Courses: A College-Level Review and Case Study}
\titlerunning{
Enhancing ML Education with HCI}
%

\author{Xiaodong Qu, Matthew Key, Eric Luo and Chuhui Qiu\\
\authorrunning{X Qu et al.}
%
\institute{
The George Washington University
\\
Washington, DC 20052 
\\
 \email{(x.qu, matthewlkey, qluo, chqiu)@gwu.edu
}
}
}
\maketitle              
\begin{abstract}
This study explores the integration of real-world machine learning (ML) projects using human-computer interfaces (HCI) datasets in college-level courses to enhance both teaching and learning experiences. Employing a comprehensive literature review, course websites analysis, and a detailed case study, the research identifies best practices for incorporating HCI datasets into project-based ML education. Key findings demonstrate increased student engagement, motivation, and skill development through hands-on projects, while instructors benefit from effective tools for teaching complex concepts. The study also addresses challenges such as data complexity and resource allocation, offering recommendations for future improvements. These insights provide a valuable framework for educators aiming to bridge the gap between theoretical knowledge and practical application in ML education.

\end{abstract}
%
%
%

\section{Introduction}
\subsection{Motivation}

The rapid advancement of Machine Learning (ML) and Artificial Intelligence (AI) technologies has created a pressing demand for skilled professionals in various industries \cite{lao2020reorienting,zheng2024charting,ng2023review,alfredo2024human,tan2024seat}. In response, higher education institutions have expanded their ML curricula to include diverse courses aimed at imparting AI literacy and problem-solving skills \cite{bennett2017teaching,kwan2014college,martins2023findings,murungi2023empowering}. These courses cater to undergraduates and working professionals seeking to enhance their expertise.

Despite the proliferation of ML courses, significant discrepancies exist in their design and delivery. Project-based ML courses, emphasizing hands-on experience and practical application, show considerable variation in structure and content, creating challenges in achieving consistent educational outcomes.

A critical aspect is the integration of datasets for practical exercises. Human-Computer Interaction (HCI) datasets offer rich opportunities for applying ML techniques. However, effective use requires careful course design and delivery.

This paper addresses key issues by conducting a comprehensive literature review, analyzing related course websites, and performing a detailed case study to identify best practices and provide actionable insights for educators. The primary audience includes computer science professors and students. The focus on learners' engagement and instructors' perspectives aims to bridge the gap between theoretical knowledge and practical application, enhancing ML education for the AI-driven job market.

\subsection{Research Questions}

The following research questions guide this study and form the basis for its analysis and conclusions:
\begin{itemize}
\item \textbf{What best practices can be identified from a review of academic papers related to project-based ML course delivery, particularly those incorporating HCI datasets?}
\item \textbf{What trends in course design, including the integration of math and statistics, and dataset usage emerge from an analysis of in-person ML courses at colleges and universities?}
\item \textbf{What insights can be gained from a detailed case study of teaching ML courses with specific HCI datasets over the past three years, considering both learner engagement and instructor perspectives?}
\end{itemize}

By addressing these questions, this paper aims to bridge the gap between theoretical knowledge and practical application in ML education. The findings are intended to guide educators in enhancing the design and delivery of project-based ML courses, ultimately improving student learning outcomes and better preparing them for the demands of the AI-driven job market.

\section{Related Works}
\subsection{Overview}
Historically, higher education in technical fields has relied on a blend of lectures, exercises, and practical labs to foster both theoretical understanding and hands-on proficiency with field-relevant methods \cite{abood2019learning}. In the domain of computer science, it is crucial for graduates to acquire practical skills that align with industrial demands, ensuring they are well-prepared for the workforce \cite{beckman1997collaborations,shaw2005deciding}. Institutions such as universities of applied sciences emphasize this alignment through curricula designed to bridge academic knowledge and practical application.

Despite these efforts, challenges persist when graduates encounter large, complex real-world projects or when they lack essential teamwork skills \cite{beckman1997collaborations,shaw2005deciding}. These challenges underscore the growing importance of project-based learning (PBL), which involves industrial topics and aims to better prepare students for professional environments \cite{winzker2012semester,daun2014industrial,daun2016project}. PBL provides students with opportunities to engage in collaborative, hands-on projects that simulate real-world scenarios, thereby enhancing their problem-solving abilities and teamwork competencies.

\subsection{Project-Based Learning in ML Education}
In rapidly evolving fields such as deep learning, a subset of Machine Learning (ML), there is a continuous emergence of new trends in algorithms, datasets, and pedagogical approaches \cite{huang2019integrating,miller2019promoting,brungel2020project,wong2020project}. Project-based learning (PBL) has been recognized as an effective pedagogical strategy in these contexts. It facilitates active learning by immersing students in real-world problems and encouraging them to apply theoretical concepts to practical challenges.

Several studies have highlighted the benefits of integrating PBL into ML education. For instance, Huang et al. (2019) emphasize the importance of incorporating contemporary datasets and real-world applications to keep the curriculum relevant and engaging \cite{huang2019integrating}. Miller et al. (2019) discuss how PBL can promote deeper understanding and retention of complex ML concepts by enabling students to work on projects that mirror industrial applications \cite{miller2019promoting}. Brungel et al. (2020) and Wong et al. (2020) further elaborate on how PBL fosters critical thinking and problem-solving skills, which are essential for success in the fast-paced field of ML \cite{brungel2020project,wong2020project}.

\subsection{K-12 ML Education}
The burgeoning interest in K-12 Machine Learning education has prompted a synthesis of existing research to better understand how ML can be effectively integrated into early education. Several studies have explored resources for ML education at the K-12 level, its integration into existing curricula, and innovative pedagogical strategies \cite{sanusi2020pedagogies,marques2020teaching,reddi2021widening,sanusi2023systematic,van2023emerging}.

Sanusi et al. (2020) and Marques et al. (2020) investigate various pedagogical approaches that can make ML concepts accessible to younger students, emphasizing the importance of foundational understanding and engagement \cite{sanusi2020pedagogies,marques2020teaching}. Reddi et al. (2021) highlight the need for curriculum development that not only introduces ML concepts but also integrates them seamlessly into subjects already being taught, thereby enriching the overall educational experience \cite{reddi2021widening}. Sanusi (2023) and Van (2023) provide systematic reviews and emerging trends in K-12 ML education, offering insights into the most effective practices and resources for educators \cite{sanusi2023systematic,van2023emerging}.

These studies collectively underscore the critical role of early ML education in fostering future generations of tech-savvy individuals who are well-prepared for advanced studies and careers in AI and ML.

\section{Methods}
 
This study employed a comprehensive review of academic papers on ML-related courses, supplemented by an extensive survey of relevant course websites. Additionally, a reflective analysis of teaching experiences over the past three years was presented as a case study.

\begin{table}[b!]
\caption{Progression of Paper Search Steps: S1 represents initial search results, S2 indicates potentially relevant findings, S3 highlights confirmed relevant results, and S4 enumerates those results after removing duplicates.}
\resizebox{\textwidth}{!}{ 
\begin{tabular}{|l|l|l|l|l|}
\hline
Paper Source (Steps)       & S1 & S2 & S3 & S4 \\\hline
Google Scholar      & 312                    & 230                                 & 112                     & 112                                          \\
ACM DIgital Library & 120                    & 90                                  & 45                      & 40                                           \\
IEEE Xplore         & 66                     & 51                                  & 37                      & 35                                           \\
ERIC                & 31                     & 21                                  & 15                      & 12                                           \\
arXiv               & 15                     & 7                                  & 5                       & 5                                            \\\hline
subtotal            & 544                    & 399                                 & 214                     & 204    \\\hline                                     
\end{tabular}
} 
\label{table:1}
\end{table}

\subsection{Keywords}

\begin{figure}[hb!]
  \centering
  \includegraphics[width=\linewidth]{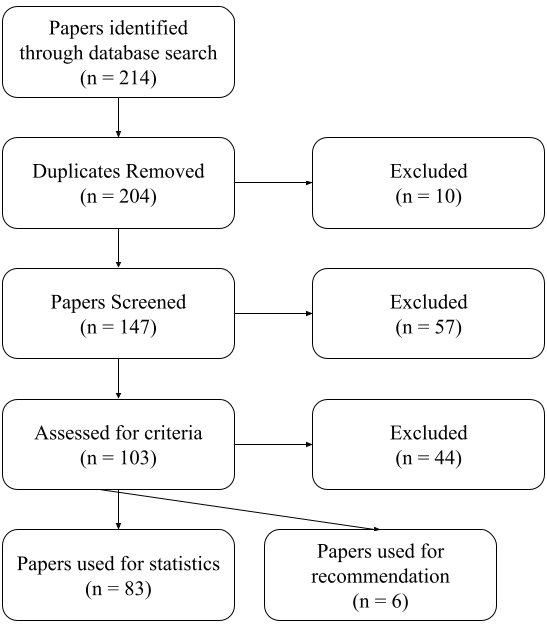} 
  \caption{Selection process for the papers}
  \label{fig:method}
\end{figure}

Utilizing the Preferred Reporting Items for Systematic Reviews and Meta-Analyses (PRISMA) approach, pertinent papers were systematically identified over a two-month period, from August to October 2023. The databases explored included Google Scholar, IEEE Xplore, ACM Digital Library, arXiv, and ERIC. The keyword search comprised: (\textit{'Machine Learning' OR 'Deep Learning' OR 'ML' OR 'DL') AND ('project-based' OR 'project-based learning' OR 'PBL') AND ('Survey' OR 'Review' OR 'Case Study')}.

Additionally, the search strings used were: \textit{"project-based learning in machine learning", "applied machine learning course design", "teaching strategies for machine learning", "HCI datasets in project-based learning", "challenges in project-based machine learning instruction", "student engagement in project-based machine learning courses"}.

This strategy aimed to pinpoint papers aligning with the research questions. Table \ref{table:1} and Figure \ref{fig:method}
 visualize the search trajectory, showcasing the number of papers identified and excluded based on set criteria. To cater to the target audience's time constraints, a concise list of papers encapsulating the prevailing trends in the domain was curated.

\subsection{Selection Criteria}

To ensure the relevance and quality of the review's content, the following criteria were applied:
\begin{itemize}
\item \textbf{Project-based:} Papers and courses must emphasize project-based machine learning, detailing their design and execution. This focus ensures that the content is practically relevant and applicable to real-world ML problems.
\item \textbf{Publication Time Frame:} Papers published from 2017 onwards and courses updated after 2020 were included. This time frame ensures that the review encompasses the most recent advancements and trends in ML education.
\item \textbf{Machine Learning Focus:} Preference was given to content primarily addressing project-based Machine Learning or Deep Learning. This focus aligns with the goal of enhancing practical ML education for professionals.
\item \textbf{Target Audience:} Papers and courses should cater to computer science (CS) professors, students, or non-CS majors enrolled in such courses. This criterion ensures that the content is relevant to educators and learners who are directly involved in ML education, as well as those from diverse academic backgrounds seeking to gain practical ML skills.
\end{itemize}

\subsection{Case Study}

Over the past three years (2022 to 2024), four project-based machine learning courses were delivered by an assistant professor in computer science at two different institutions. In 2021 and 2022, undergraduate students at Swarthmore College, PA, participated in these courses. In 2023, graduate students at George Washington University, DC, were enrolled in similar project-based machine learning courses. These courses aimed to evaluate the effectiveness of project-based learning in enhancing students' practical machine learning skills and their ability to apply theoretical knowledge to real-world problems. Detailed course outlines, project descriptions, and instructional methodologies were documented and analyzed. These materials are available on the faculty page of the instructor \footnote{https://faculty.cs.gwu.edu/xiaodongqu/}.

A 2021 EEG dataset was utilized in these machine learning courses. Electroencephalography (EEG) is extensively used in research fields such as neural engineering, neuroscience, biomedical engineering, and brain-like computing, with particular emphasis on brain-computer interfaces (BCIs). The analysis of EEG signals is crucial for the development of BCIs, providing deep insights into the complex neural activities of the human brain. Over the past decade, a variety of machine learning and deep learning algorithms have been applied to EEG data, leading to significant advancements in numerous applications. These applications include emotion recognition, motor imagery, mental workload assessment, seizure detection, Alzheimer's disease classification, sleep stage scoring, and many more \cite{kastrati2021eegeyenet,craik2019deep,roy2019deep,altaheri2023deep,qu2022time,gao2021complex,hossain2023status,yi2022attention,key2024advancing,li2024enhancing,koome2023trends,murungi2023empowering,dou2022time,zhou2022brainactivity1,qu2020identifying,qu2020using,qu2020multi,qu2018eeg,qu2019personalized,saeidi2021neural,qu2022eeg4home,rasheed2020machine,dadebayev2022eeg,wang2022eeg,li2020deep,aggarwal2022review}.

This case study used a mixed-methods approach to gather comprehensive feedback from students. Thirty-five students were interviewed after completing the courses, providing qualitative data on their learning experiences, challenges, and perceived benefits. Additionally, anonymous course feedback forms were collected and analyzed, offering quantitative insights into student satisfaction and instructional effectiveness.

Key aspects highlighted by student feedback include:
\begin{itemize}
\item \textbf{Engagement and Motivation:} High levels of engagement and motivation due to the hands-on nature of the projects, showing practical applications of ML concepts.
\item \textbf{Skill Development:} Significant improvements in technical skills, especially in data preprocessing, model development, and evaluation, with real-world datasets enhancing problem-solving abilities.
\item \textbf{Collaboration and Teamwork:} Encouraged collaboration, fostering a collaborative learning environment, and developing essential teamwork skills.
\item \textbf{Challenges and Areas for Improvement:} Challenges with project complexity and the steep learning curve of advanced ML techniques. Suggestions included more support during initial project stages.
\end{itemize}

The analysis of course feedback revealed that the project-based learning approach was well-received, with students valuing the practical application of theoretical knowledge. Insights from this case study inform recommendations for enhancing project-based ML courses, making them more effective and accessible for students at various academic levels.

This case study underscores the importance of integrating practical, hands-on projects in ML education. It provides valuable insights for educators seeking to design courses that impart theoretical knowledge and practical skills for success in the evolving field of ML.

\section{Results} 

\subsection{Literature Review}   
The literature review focused on identifying best practices in Machine Learning (ML) education, particularly in in-person courses offered by colleges and universities. The findings were categorized into several key topics, highlighting common best practices across various studies. Table \ref{table:result01} summarizes these findings.

\begin{table}[h]
\caption{Paper Results Key: U denotes undergrad-only studies, G for graduate-only, and UG for both levels. R signifies review papers, C indicates case studies, and Best P stands for best practices. }
\resizebox{\textwidth}{!}{ 
\begin{tabular}{|l|l|l|l|}
\hline
Paper                      & Level          & Type         & Best P\\ \hline
{[}1,2,4,7,15,19,20{]}  & U & R          & 1,2,3          \\ \hline
{[}5,9,12,14,17,25{]}   & U & C      & 2,3,4          \\ \hline
{[}3, 6, 11, 18, 21{]} & UG         & R           & 3,5            \\ \hline
{[}22, 23,26, 28,31{]}     & UG          & C        & 2,6            \\ \hline
{[}16, 24, 29, 33{]}       & G      & R         & 1,2,3,5        \\ \hline
{[}8, 10, 13, 16, 34{]}    & G      & C      & 1, 3, 4        \\ \hline
\end{tabular}
} 
\label{table:result01}
\end{table}

\hfill \break
\textbf{1. Machine Learning:}

\begin{itemize}
\item \textbf{Comprehensive Curriculum:} Effective ML courses offer a well-rounded curriculum covering fundamental concepts, advanced techniques, and practical applications. This ensures learners gain a broad understanding and can apply knowledge to real-world problems, benefiting instructors with a structured roadmap for teaching complex topics.
\item \textbf{Integration of Math and Statistics:} A solid foundation in mathematics and statistics is crucial for understanding ML algorithms. Effective courses include relevant topics like linear algebra, calculus, probability, and statistical inference, equipping students with essential analytical skills.
\item \textbf{Hands-On Projects:} Incorporating hands-on projects enhances ML education. Real-world datasets and practical problems significantly improve learning outcomes by providing practical experience and reinforcing theoretical knowledge, while also helping instructors assess student understanding.
\end{itemize}

\textbf{2. Project-Based Teaching and Learning:}

\begin{itemize}
\item \textbf{Engagement and Motivation:} Project-based learning (PBL) increases student engagement and motivation. Working on relevant projects keeps learners invested, leading to more dynamic and interactive classrooms.
\item \textbf{Collaborative Learning:} Successful PBL courses encourage student collaboration through group projects and peer feedback, promoting deeper understanding and teamwork skills, while easing the instructional burden.
\item \textbf{Practical Application:} PBL emphasizes applying theoretical knowledge to real-world problems, helping students develop critical thinking and problem-solving skills for their future careers.
\end{itemize}

\textbf{3. Exam-Based Teaching:}

\begin{itemize}
\item \textbf{Knowledge Assessment:} Exam-based teaching assesses students' theoretical understanding through written exams and quizzes, effectively measuring knowledge retention and application.
\item \textbf{Standardization:} Exams provide a consistent evaluation method across students, useful for large classes where individual project assessment is impractical.
\item \textbf{Individual Focus:} Exams emphasize individual performance, fostering independent study habits, though they may not capture collaborative and practical skills.
\item \textbf{Certification Preparation:} Exams prepare students for professional certification tests, which often follow similar formats, aiding in obtaining required credentials.
\end{itemize}

\textbf{4. Formal Course Structure:}

\begin{itemize}
\item \textbf{Structured Schedules:} In-person ML courses often benefit from structured schedules that ensure consistent progress and regular engagement. These schedules help students balance their academic responsibilities and maintain steady progress. Instructors can design courses with regular milestones and assessments to keep students on track.
\item \textbf{Support and Resources:} Providing ample support and resources is critical for in-person course settings. This includes access to office hours, tutoring, instructional videos, and supplemental materials that help learners overcome obstacles. For instructors, it means developing comprehensive resources that can guide students through their learning journey and provide additional support as needed.
\end{itemize}

\textbf{5. Students' Feedback:}

\begin{itemize}
\item \textbf{Positive Impact of Structured Learning:} Students frequently highlight the benefits of the structured schedule offered by in-person courses. This structure allows them to balance their studies with other academic and personal commitments effectively. Instructors receive positive feedback on course accessibility and organization, which can enhance course ratings and attract more students.
\item \textbf{Need for Interactive Elements:} Feedback often suggests that incorporating interactive elements, such as quizzes, labs, and real-time feedback, can enhance the learning experience in in-person courses. Instructors can leverage this feedback to design more engaging and interactive course content, improving student satisfaction and outcomes.
\end{itemize}

\textbf{6. Professors' Feedback:}

\begin{itemize}
\item \textbf{Importance of Course Design:} Professors emphasize the significance of well-structured course design in in-person learning environments. Clear learning objectives, organized content, and regular assessments are essential for maintaining student engagement and ensuring successful learning outcomes. Effective course design helps instructors manage course delivery more efficiently and ensures that learning goals are met.
\item \textbf{Challenges in Providing Support:} While in-person courses offer structure, professors note the challenges in providing timely support and feedback to students. Implementing systems and leveraging technology can help address these challenges. Instructors can use tools such as automated grading systems, learning management systems (LMS), and discussion forums to provide timely support.
\end{itemize}

By addressing both the learners' and instructors' perspectives and comparing project-based teaching with exam-based teaching, this section provides a comprehensive view of the best practices in ML education within in-person college and university settings. It highlights the benefits and challenges of both teaching methods, offering insights that can inform the design and implementation of effective ML courses.

\subsection{Course Websites Analysis}   

Table \ref{table:result02} showcases the findings from from the analysis of existing course websites. This analysis aimed to identify common elements and best practices in Machine Learning (ML) courses offered in-person at colleges and universities. Several key topics were explored to understand how these courses are structured and what resources they provide to both learners and instructors.

\begin{table}[ht!]
\caption{Analysis of Course Websites: U represents undergrad-only courses, G for graduate-only, and UG for both. Best P signifies courses emphasizing best practices.}
\resizebox{\textwidth}{!}{ 
\begin{tabular}{|l|l|l|l|}
\hline
Course           & Level & Institution & Best P      \\ \hline
{[}41, 42{]}     & U     & Williams       &  7,8         \\ \hline
{[}43{]}         & U     & Amherst         & 8           \\ \hline
{[}44,45{]}      & U     & Swarthmore       & 7,9,11,12         \\ \hline
{[}46, 47,48{]}  & U     & Pomona          & 7,8,9,10       \\ \hline
{[}49, 50{]}     & UG    & Harvard         & 7,8,9,11       \\ \hline
{[}50, 51{]}     & UG    & Upenn            & 7, 10, 11,12     \\ \hline
{[}52, 53, 54{]} & UG    & Stanford       & 7,8,9,10,11,12 \\ \hline
{[}55, 56{]}     & UG    & MIT            & 8,9,10,11      \\ \hline
{[}57,58,59{]}   & UG    & CMU             & 8, 9,11      \\ \hline
{[}60,61,62{]}   & UG    & UC B    & 7,8,9,10, 11 \\ \hline
\end{tabular}

} 
 
\label{table:result02}
\end{table}

\hfill \break
\textbf{7. Course Structure:}

\begin{itemize}
\item \textbf{Modular Structure:} Many analyzed courses feature a modular structure, allowing for systematic progression through the material. This structure is beneficial for organizing the curriculum in a way that builds upon foundational concepts before advancing to more complex topics. For instructors, it provides a clear framework for delivering content effectively.
\item \textbf{Integration of Math and Statistics:} Effective ML courses integrate essential mathematics and statistics topics, such as linear algebra, calculus, probability, and statistical inference. This integration is crucial for students to understand the theoretical underpinnings of ML algorithms. Instructors can use this foundation to explain complex concepts and ensure students are well-prepared for practical applications.
\end{itemize}

\hfill \break
\textbf{8. Progress Tracking:}

\begin{itemize}
\item \textbf{Tools for Tracking Progress:} Effective courses often include tools for tracking progress, such as dashboards that display completed modules and upcoming tasks. These tools help learners stay organized and motivated. For instructors, progress tracking tools provide insights into student performance and areas that may require additional attention.
\end{itemize}

\hfill \break
\textbf{9. Project-Based Teaching and Learning:}

\begin{itemize}
\item \textbf{Hands-On Projects:} A significant number of courses incorporate project-based learning, where students work on real-world projects to apply the concepts they have learned. These projects often involve datasets from industry or research, providing practical experience. For instructors, hands-on projects offer a practical way to assess students' application of theoretical knowledge.
\item \textbf{Peer Collaboration:} Some courses facilitate peer collaboration through discussion forums or group projects, allowing learners to share insights and provide mutual support. This collaboration helps build teamwork skills, which are valuable in professional settings. Instructors benefit from the collaborative learning environment as it can enhance student engagement and reduce the instructional burden.
\end{itemize}

\hfill \break
\textbf{10. Sample Code:}

\begin{itemize}
\item \textbf{Code Repositories:} Many courses provide access to code repositories, such as GitHub, where learners can find sample code and scripts used in the course. This is particularly useful for understanding practical implementation details. Instructors can use these repositories to demonstrate coding practices and provide students with resources for independent study.
\item \textbf{Code Walkthroughs:} Courses that include detailed code walkthroughs, either in written form or through video demonstrations, help learners understand the step-by-step process of developing ML models. For instructors, code walkthroughs are an effective teaching tool to illustrate coding techniques and problem-solving strategies.
\end{itemize}

\hfill \break
\textbf{11. Lecture Slides:}

\begin{itemize}
\item \textbf{Comprehensive Lecture Slides:} High-quality courses offer comprehensive lecture slides that summarize key concepts and provide visual aids to enhance understanding. These slides are often available for download, allowing learners to review them at their own pace. Instructors can use these slides to structure their lectures and provide students with a consistent reference material.
\item \textbf{Supplemental Materials:} In addition to slides, some courses provide supplemental materials such as cheat sheets, reference guides, and additional readings to deepen learners' understanding. These materials support instructors in offering a richer educational experience and cater to diverse learning needs.
\end{itemize}

\hfill \break
\textbf{12. Course Videos:}

\begin{itemize}
\item \textbf{Engaging Video Lectures:} Video lectures are a staple of in-person ML courses, often used to supplement classroom teaching. The best courses feature engaging, well-produced videos that clearly explain complex concepts. These videos often include demonstrations, animations, and real-world examples to illustrate key points. Instructors can leverage these videos to reinforce classroom teaching and provide students with additional learning resources.
\item \textbf{Interactive Elements:} Some courses incorporate interactive elements within videos, such as embedded quizzes or coding challenges, to reinforce learning and keep learners engaged. These elements provide immediate feedback to students and help instructors gauge understanding in real-time.
\end{itemize}

By addressing both the learners' and instructors' perspectives, this section provides a comprehensive view of the best practices in ML education within in-person college and university settings. It highlights the importance of integrating foundational math and statistics, hands-on projects, and collaborative learning, offering insights that can inform the design and implementation of effective ML courses.

\subsection{Case Study}   
The case study examines the implementation of project-based learning in Machine Learning (ML) courses from 2022 to 2024 at Swarthmore College and George Washington University. These courses provided students with hands-on experience applying ML techniques to real-world problems using a brain-computer interfaces (BCI) dataset.

\textbf{Course Context and Structure:}

Courses included lectures, hands-on projects, and collaborative activities to bridge theoretical knowledge and practical application, ensuring students could apply ML concepts to complex problems.

\begin{itemize}
\item \textbf{Instructors' Perspective:} Designed by an assistant professor in computer science to create a challenging yet supportive environment, the courses used a BCI dataset for advanced ML applications in Human-Computer Interaction (HCI).
\item \textbf{Learners' Perspective:} Undergraduate and graduate students engaged deeply with material, developing practical skills highly valued in the industry.
\end{itemize}

\textbf{Implementation of the BCI Dataset:}

The BCI dataset consisted of EEG recordings from subjects engaged in tasks. This dataset's complexity and relevance to HCI research made it an ideal choice.

\begin{itemize}
\item \textbf{Project Design:} Students developed ML models to classify mental states based on EEG data, involving preprocessing, feature extraction, and model training and evaluation.
\item \textbf{Instructors' Role:} Instructors guided technical aspects, facilitated group discussions, and provided feedback.
\item \textbf{Learners' Experience:} Students reported high engagement and motivation, finding the project a valuable learning experience that challenged them to think critically and innovate.
\end{itemize}

\textbf{Outcomes and Feedback:}

Course outcomes were evaluated through student feedback, project assessments, and instructor observations.

\begin{itemize}
\item \textbf{Engagement and Motivation:} Students showed increased engagement and deeper interest in HCI and ML applications, citing the challenging dataset as a key factor.
\item \textbf{Skill Development:} Both instructors and students noted significant improvements in technical skills, particularly in data preprocessing, feature extraction, and model development.
\item \textbf{Collaboration and Teamwork:} Courses facilitated collaborative learning, enhancing understanding of ML concepts and developing teamwork and communication skills.
\item \textbf{Instructor Insights:} Instructors found the project-based approach effective in promoting deep learning and practical skill acquisition, with the BCI dataset enriching the experience.
\end{itemize}

\textbf{Challenges and Recommendations:}

Several challenges were encountered during the courses.

\begin{itemize}
\item \textbf{Data Complexity:} The BCI dataset's complexity challenged students, especially those with limited ML experience. Instructors provided additional support and resources, such as tutorials on EEG data processing.
\item \textbf{Time Management:} Balancing project workload with other requirements was challenging. Instructors recommended clearer guidelines and structured timelines.
\item \textbf{Resource Allocation:} The resource-intensive nature of project-based learning required significant instructor time. Future courses could benefit from more teaching assistants or automated tools.
\end{itemize}

In conclusion, the case study highlights the effectiveness of project-based learning in ML education, particularly using complex datasets like BCI. Both learners and instructors benefited from the hands-on, collaborative approach, facilitating deep learning and practical skill development. These insights provide valuable guidance for educators implementing similar approaches.

\section{Discussion}
 
\subsection{Real-World Machine Learning Projects with HCI Datasets}

A significant finding from this study is the crucial role of real-world machine learning projects with HCI datasets in enhancing both teaching and learning experiences. These projects provide learners with hands-on experience and practical application of the concepts they have learned. Instructors also benefit from these projects as they provide a rich context for teaching complex ML topics and assessing student understanding. Here is a subset of the experiments that learners and instructors have explored and practiced so far \cite{an2023transfer,an2023survey,li2024spherehead,chen2024hytrel,chen2021explicitly,lu2023machine,lu2024uncertainty,lu2023deep,zhao2024deep,zhao2022uda,wang2023robust,tang2023active,qu2020multi,qu2020using,gui2024remote,yunoki2023exploring,dou2022time,tan2023state,murungi2023empowering,zhang2022attention,yunoki2023exploring,zhang2023trep,yi2022attention,ma2022traffic,ma2024data,jiang2023successfully}.

The projects discussed involved various HCI datasets, including brain-computer interfaces (BCI), eye-tracking, and gesture recognition data. These datasets were chosen for their relevance to cutting-edge HCI research and their ability to challenge students to apply ML techniques to real-world problems.

\subsubsection{Learners' Perspective}

Working with HCI datasets offers several benefits:

\begin{itemize}
\item \textbf{Engagement and Motivation:} Real-world projects significantly increase engagement and motivation, showing the relevance of studies.
\item \textbf{Skill Development:} Students develop critical skills like data preprocessing, feature extraction, and model development.
\item \textbf{Critical Thinking and Problem-Solving:} Projects require critical thinking and innovative solutions, building problem-solving skills.
\item \textbf{Collaboration and Teamwork:} Collaborative projects foster teamwork and communication skills.
\end{itemize}

A student noted that applied ML courses felt like multiple courses in one, demanding self-guided learning in problem selection, literature review, proposing novel methods, and technical implementation. This comprehensive approach significantly contributed to their skill development.

\subsubsection{Instructors' Perspective}

Real-world projects with HCI datasets offer several advantages:

\begin{itemize}
\item \textbf{Effective Teaching Tool:} HCI datasets make abstract ML concepts concrete and understandable.
\item \textbf{Assessment of Student Understanding:} Projects provide a practical means of assessing ML concept application.
\item \textbf{Enhanced Engagement and Interaction:} Projects lead to dynamic, interactive classrooms with high participation.
\item \textbf{Resource for Research and Development:} Student projects can contribute to ongoing research, exploring new ideas.
\end{itemize}

\subsubsection{Examples of HCI Projects}

Examples of projects include:

\begin{itemize}
\item BCI: Developing models to classify mental states based on EEG signals.
\item Eye-Tracking: Predicting user intent and analyzing gaze patterns to improve UI design.
\item Gesture Recognition: Creating models to recognize and interpret human gestures for device control.
\end{itemize}

These projects provided insights into ML applications in HCI, offering a deeper understanding of the technologies' potential and challenges.

\subsubsection{Challenges and Recommendations}

Challenges encountered:

\begin{itemize}
\item \textbf{Data Complexity:} HCI datasets can be overwhelming. Instructors should provide tutorials and support materials.
\item \textbf{Time Management:} Balancing project work with other requirements is challenging. Clear guidelines and structured timelines are essential.
\item \textbf{Resource Allocation:} Project-based learning requires significant instructor time. Additional teaching assistants or automated tools can help manage the workload.
\end{itemize}

\subsection{Future Work}

Future work can build on this study's findings:

\begin{itemize}
\item \textbf{Expanding Dataset Variety:} Incorporate more HCI datasets, including speech recognition, NLP, and VR interactions, to provide broader exposure.
\item \textbf{Enhancing Support for Novice Researchers:} Develop structured support for students with little research experience, such as mentorship programs, research guides, and workshops.
\item \textbf{Leveraging Technology for Support:} Utilize AI-driven tutoring and automated feedback tools to provide timely, personalized support, especially in large classes.
\item \textbf{Longitudinal Studies on Learning Outcomes:} Conduct longitudinal studies to track the long-term impact of project-based learning on careers and research outputs.
\item \textbf{Integrating Interdisciplinary Approaches:} Encourage interdisciplinary projects combining ML with fields like psychology, neuroscience, and engineering for innovative solutions.
\item \textbf{Comparative Studies of ML Techniques:} Explore various machine learning techniques on different datasets for comparative studies \cite{an2023transfer,an2023survey,jiang2023successfully,lu2023machine,chen2024trialbench,ma2022traffic,ma2024data,gui2024remote,tan2023audio,tan2021multivariate,qiu2023modal,zhao2024deep} to provide valuable insights and further enhance the field.
\end{itemize}

In conclusion, integrating real-world ML projects with HCI datasets enhances teaching and learning. These projects provide practical skills and critical thinking abilities, while offering instructors effective tools for teaching complex concepts and assessing understanding. The insights from this study guide educators in implementing similar approaches.

\section{Conclusion} 

This study highlights the effectiveness of integrating real-world machine learning (ML) projects with human-computer interfaces (HCI) datasets in enhancing both teaching and learning experiences. By providing hands-on experience and practical application opportunities, these projects significantly increase student engagement, motivation, and skill development. From the instructors' perspective, they offer valuable tools for teaching complex concepts and assessing student understanding. Despite the challenges of data complexity and resource allocation, the benefits of project-based learning are evident. Future work should focus on expanding dataset variety, enhancing support for novice researchers, leveraging technology for support, and conducting longitudinal studies on learning outcomes. The insights gained from this study provide a robust framework for educators seeking to implement similar approaches, ultimately bridging the gap between theoretical knowledge and practical application in ML education.

%
%
%
 \bibliographystyle{splncs04}
 \bibliography{teach_ML}

\end{document}